\pdfoutput=1
\documentclass{article}
\usepackage{ijcai20}

\usepackage{times}
\usepackage{soul}
\usepackage[hidelinks]{hyperref}
\usepackage[utf8]{inputenc}
\usepackage[small]{caption}
\usepackage{graphicx}
\usepackage{amsmath}
\usepackage{amsthm}
\usepackage{booktabs}
\usepackage{algorithm}
\usepackage{algorithmic}
\usepackage{balance}
\usepackage{multirow}
\usepackage{subfig}
\usepackage{diagbox}

\title{AdaBERT: Task-Adaptive BERT Compression with Differentiable \\ Neural Architecture Search}

\author{
Daoyuan Chen \thanks{Equal contribution.},\hspace{1mm} Yaliang Li $^*$,\hspace{1mm} Minghui Qiu,\hspace{1mm} Zhen Wang,\hspace{1mm} Bofang Li,\hspace{1mm}\\
Bolin Ding, \hspace{1mm} Hongbo Deng,\hspace{1mm} Jun Huang,\hspace{1mm} Wei Lin,\hspace{1mm} Jingren Zhou\\
\affiliations
Alibaba Group\\
\emails
\{daoyuanchen.cdy, yaliang.li, minghui.qmh, jones.wz, bofang.lbf\}@alibaba-inc.com \\
\{bolin.ding, dhb167148, huangjun.hj, weilin.lw, jingren.zhou\}@alibaba-inc.com
}

\begin{document}
\maketitle
\begin{abstract}
Large pre-trained language models such as BERT have shown their effectiveness in various natural language processing tasks. 
However, the huge parameter size makes them difficult to be deployed in real-time applications that require quick inference with limited resources. 
Existing methods compress BERT into small models while such compression is task-independent, i.e., the same compressed BERT for all different downstream tasks. 
Motivated by the necessity and benefits of task-oriented BERT compression, we propose a novel compression method, AdaBERT, that leverages differentiable Neural Architecture Search to automatically compress BERT into task-adaptive small models for specific tasks. 
We incorporate a task-oriented knowledge distillation loss to provide search hints and an efficiency-aware loss as search constraints, which enables a good trade-off between efficiency and effectiveness for task-adaptive BERT compression. 
We evaluate AdaBERT on several NLP tasks, and the results demonstrate that those task-adaptive compressed models are 12.7x to 29.3x faster than BERT in inference time and 11.5x to 17.0x smaller in terms of parameter size, while comparable performance is maintained.
\end{abstract}

\section{Introduction}
\label{sec:intro}

Nowadays, pre-trained contextual representation encoders, such as ELMo \cite{peters2018deep}, BERT and \cite{devlin2019bert}, and RoBERTa \cite{liu2019roberta}, have been widely adopted in a variety of Natural Language Processing (NLP) tasks. 
Despite their effectiveness, these models are built upon large-scale datasets and they usually have parameters in the billion scale. For example, the BERT-base and BERT-large models are with $109$M and $340$M parameters respectively. 
It is difficult to deploy such large-scale models in real-time applications that have tight constraints on computation resource and inference time.

To fulfill the deployment in real-time applications, recent studies compress BERT into a relatively small model to reduce the computational workload and accelerate the inference time.
BERT-PKD \cite{sun2019patient} distills BERT into a small Transformer-based model that mimics intermediate layers from original BERT. 
TinyBERT \cite{jiao2019tinybert} uses two-stage knowledge distillation and mimics attention matrices and embedding matrices of BERT.
And in \cite{Michel2019aresixteen}, the authors propose a method to iteratively prune the redundant attention heads of BERT.

However, these existing studies compress BERT into a task-independent model structure, i.e., the same compressed BERT structure for all different tasks.
Recall that BERT learns various knowledge from the large-scale corpus, while only certain parts of the learned knowledge are needed for a specific downstream task \cite{tenney2019bertRedis}. 
Further, \cite{liu2019linguistic} show that different hidden layers of BERT learned different levels of linguistic knowledge, and \cite{clark2019does} demonstrates that the importance degrees of attention heads in BERT vary for different tasks. 
All these findings shed light on the task-adaptive BERT compression: different NLP tasks use BERT in different ways, and it is necessary to compress large-scale models such as BERT for specific downstream tasks respectively.
By doing so, the task-adaptive compressed BERT can remove task-specific redundant parts in original large-scale BERT, which leads to better compression and faster inference.

Motivated by this, we propose a novel \textbf{Ada}ptive \textbf{BERT} compression method, \textit{AdaBERT}, that leverages differentiable Neural Architecture Search (NAS) to automatically compress BERT into task-adaptive small models for specific tasks. 
We incorporate a task-oriented knowledge distillation loss that depends on the original BERT model to provide search hints, as well as an efficiency-aware loss based on network structure as search constraints. These two loss terms work together and enable the proposed compression method to achieve a good trade-off between efficiency and effectiveness for different downstream tasks.
To be specific, we adopt a lightweight CNN-based search space and explicitly model the efficiency metrics with respect to searched architectures, which has not been considered in previous BERT compression studies.
Further, we hierarchically decompose the learned general knowledge of BERT into task-oriented useful knowledge with a set of probe models for knowledge distillation loss, such that the architecture search space can be reduced into a small task-oriented sub-space.
Finally, by relaxing the discrete architecture parameters into continuous distribution, the proposed method can efficiently find task-adaptive compression structures through the gradient-based optimization.

To evaluate the proposed method, we compress BERT for several NLP tasks including sentiment classification, entailment recognition, and semantic equivalence classification.
Empirical results on six datasets show that the proposed compression method can find task-adaptive compression models that are 12.7x to 29.3x faster than BERT in inference time and 11.5x to 17.0x smaller than BERT in terms of parameter size while maintaining comparable performance.

\section{Methodology}
\label{sec:method}

\subsection{Overview}
\begin{figure}[tp]
\centering
\includegraphics[width=0.99\linewidth]{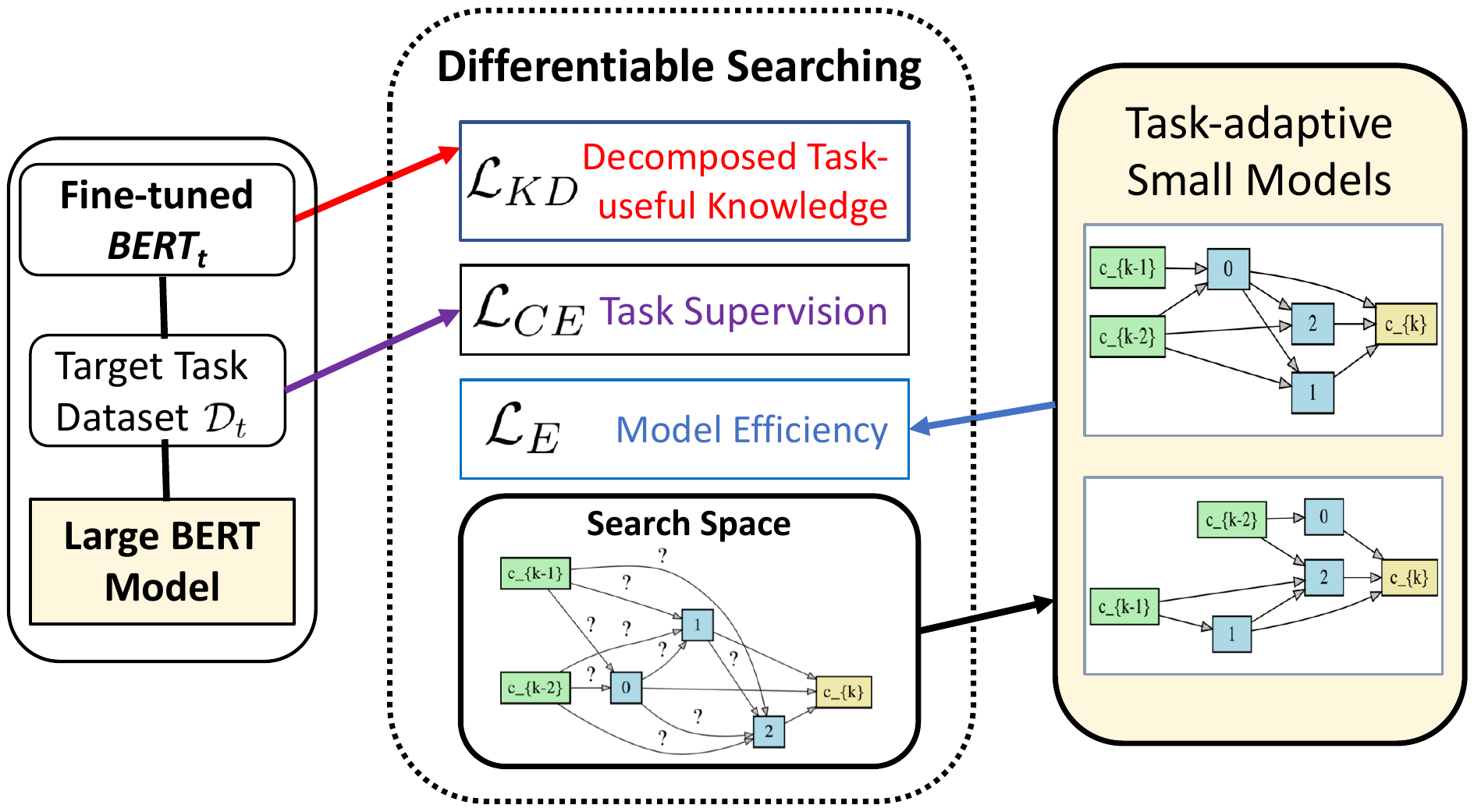}
\caption{The overview of AdaBERT.}
\label{fig:overview}
\end{figure}

As shown in Figure \ref{fig:overview}, we aim to compress a given large BERT model into an effective and efficient task-adaptive model for a specific task. The structures of the compressed models are searched in a differentiable manner with the help of task-oriented knowledge from the large BERT model while taking the model efficiency into consideration.

Formally, let's denote a BERT model fine-tuned on the target data $\mathcal{D}_t$ as $BERT_t$, an architecture searching space as $\mathcal{A}$. Our task is to find an optimal architecture $\alpha \in \mathcal{A}$ by minimizing the following loss function:

\begin{equation}
\label{equ:overall-loss}
\begin{aligned}
    \mathcal{L} = & (1-\gamma) \mathcal{L}_{CE}(\alpha, w_\alpha, \mathcal{D}_t)~ + \\ &\gamma \mathcal{L}_{KD}(\alpha, w_\alpha, BERT_t) + \beta\mathcal{L}_{E}(\alpha),
\end{aligned}
\end{equation}
where $w_\alpha$ is the trainable network weights of the architecture $\alpha$ (e.g., weights of a feed forward layer), $\mathcal{L}_{CE}$, $\mathcal{L}_{KD}$, and $\mathcal{L}_E$ are losses for the target task, task-oriented knowledge distillation and efficiency respectively. Specifically, $\mathcal{L}_{CE}$ is the cross-entropy loss w.r.t. labels from the target data $\mathcal{D}_t$, $\mathcal{L}_{KD}$ is the task-oriented knowledge distillation (KD) loss that provides hints to find suitable structures for the task, and $\mathcal{L}_E$ is the efficiency-aware term to provide constraints to help search lightweight and efficient structures. $\gamma$ and $\beta$ are hyper-parameters to balance these loss terms.

\subsection{Search Space $\mathcal{A}$}
\label{sec:space}
Most neural architecture search methods focus on cell-based micro search space \cite{zoph2017neural,pham2018efficient,liu2019darts}. 
That is, the searching target is a cell and the network architecture is stacked by the searched cell over pre-defined $K_{max}$ layers, where the cell structure parameter $\alpha_c$ is shared for all layers.
In this work, we consider a \textit{macro search space} over the entire network to enhance the structure exploration. 
Specifically, besides for searching shared cell parameter $\alpha_c$ for stacking, we also search the number of stacking layers $K \in [1, 2, \dots, K_{max}]$.
The $K$ is crucial for finding a trade-off between model expressiveness and efficiency, as a larger $K$ leads to higher model capacity but slower inference. 

\begin{figure}[]
\centering
\includegraphics[width=0.64\linewidth]{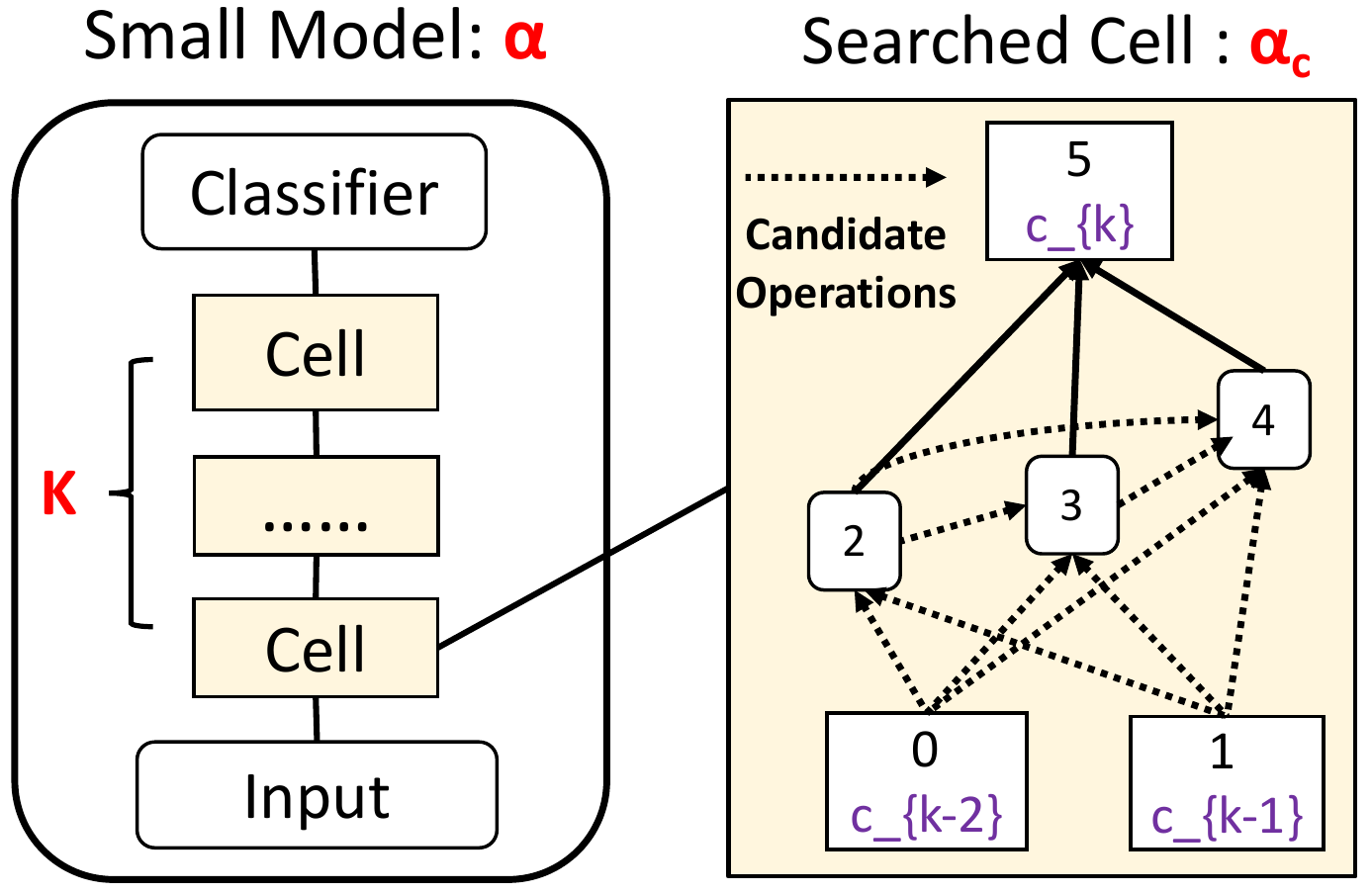}
\caption{Search space including stacked layers and stacked cells.}
\label{fig:search-space}
\end{figure}

As depicted in Figure \ref{fig:search-space}, the searched cell is represented as a directed acyclic graph. 
Each node within the cell indicates a latent state $h$ and the edge from node $i$ to node $j$ indicates operation $o_{i,j}$ that transforms $h_i$ to $h_j$.
For the cell at $k$-th layer ($k > 1$), we define two input nodes $c_{k-2}$ and $c_{k-1}$ as layer-wise residual connections, and an output node $c_{k}$ that is obtained by attentively summarized over all the intermediate nodes.
For the cell at first layer, node $0$ and node $1$ are task-dependent input embeddings.
Formally, let's denote $\mathcal{O}$ as the set of candidate operations. We assume a topological order among $N$ intermediate nodes, i.e., $o_{i,j} \in \mathcal{O}$ exists when $i < j$ and $j > 1$, and the search space $\alpha$ can be formalized as:
\begin{equation}
\begin{aligned}
    \alpha &= \{K, \alpha_c\}, ~~  K \leq K_{max}, \\
   \alpha_c &=[o_{0,2}, o_{1,2}, \dots, o_{i,j}, \dots, 
   o_{N+1,N+2}].
\end{aligned}
\end{equation}

\subsection{Task-oriented Knowledge Distillation}
\label{sec:knowledge loss}
To encourage the learned structure to be suitable for the target task, we introduce the task-oriented knowledge distillation loss, denoted as $\mathcal{L}_{KD}$ in Equation (\ref{equ:overall-loss}), to guide the structure search process.

\paragraph{Task-useful Knowledge Probe.}
We leverage a set of probe classifiers to hierarchically decompose the task-useful knowledge from the teacher model $BERT_t$, and then distill the knowledge into the compressed model. 
Specifically, we freeze the parameters of $BERT_t$, and train a Softmax probe classifier for each hidden layer w.r.t. the ground-truth task labels. In total we have $J$ classifiers, ($J=12$ in BERT-base), and the classification logits of $j$-th classifier can be regarded as the learned knowledge from $j$-th layer.
Given an input instance $m$, denote $C_{j,m}^T$ as the hidden representation from the $j$-th layer of $BERT_t$, $C_{i,m}^S$ as the attentively summed hidden state on $i$-th layer of the compressed student model,
we distill the task-useful knowledge (classification logits) as:
\begin{equation}
    \mathcal{L}_{KD}^{i,m} = -P^T_j(C_{j,m}^T) \cdot log(P^S_i(C_{i,m}^S)/T),
\end{equation}
where $T$ is the temperature value,  $P^T_j$ is the $j$-th teacher probe classifier, $P^S_i$ is the trainable student probe on $i$-th layer of the compressed model. Clearly ${L}_{KD}^{i,m}$ represents the mastery degree of teacher knowledge from the $i$-th layer of the compressed model, i.e., how similar are the logits of the two models predict for the same input $m$.
Here we set $j$ to be proportional to the layers of the two models, i.e., $j = \lceil i\times J/K \rceil $, such that the compressed model can learn the decomposed knowledge smoothly and hierarchically.

\paragraph{Attentive Hierarchical Transfer.}
The usefulness of each layer of BERT is varied for different tasks as shown in \cite{liu2019linguistic}.
Here we attentively combine the decomposed knowledge for all layers as:
\begin{equation}
\begin{aligned}
        \mathcal{L}_{KD} &= \sum_{m=0}^{M}\sum_{i=1}^{K} w_{i,m} \cdot \mathcal{L}_{KD}^{i,m}, \\ 
        w_{i,m} &= \frac{\exp[y_m \cdot logP^T_j(C_{j,m}^S)]}{\sum_{i'} \exp[y_m \cdot logP^T_{j'}(C_{j',m}^S)]},
\end{aligned}
\end{equation}
where $M$ is the total number of training instances, $y_m$ is the label of instance $m$. $w_{i,m}$ is set as the normalized weight according to the negative cross-entropy loss of the teacher probe $P^T_j$, so that probe classifiers making preciser predictions (smaller loss) gain higher weights. 
Besides, to enrich task-useful knowledge, we perform data augmentation on target task datasets with the augmentation process used in \cite{jiao2019tinybert}, which uses BERT to replace original words.

\subsection{Efficiency-Aware Loss}
\label{sec:efficiency loss}
Recall that we aim to compress the original BERT model into efficient compressed models. To achieve this, we further incorporate model efficiency into loss function from two aspects, i.e., parameter size and inference time.
To be specific, for searched architecture $\alpha_c$ and $K$, we define the efficiency-aware loss in Equation (\ref{equ:overall-loss}) as:
\begin{equation}
\begin{aligned}
        \mathcal{L}_E = \frac{K}{K_{max}} \sum_{o_{i,j} \in \alpha_c}
        SIZE(o_{i,j})+FLOPs(o_{i,j}),
        \end{aligned}
\end{equation}
where $K_{max}$ is the pre-defined maximum number of layers, $SIZE(\cdot)$ and $FLOPs(\cdot)$ are the normalized parameter size and the number of floating point operations (FLOPs) for each operation. 
The sum of FLOPs of searched operations serves as an approximation to the actual inference time of the compressed model.

\subsection{Differentiable Architecture Searching}
\label{sec:search algo}

A major difference between the proposed method and existing BERT compression methods is that the proposed AdaBERT method seeks to find task-adaptive structures for different tasks. Now we will discuss the task-adaptive structure searching via a differentiable structure search method with the aforementioned loss terms.

\subsubsection{Search Space Setting}
Before diving into the details of the search method, we first illustrate the search space for our method in Figure~\ref{fig:search-space}. To make it easy to stack cells and search for network layers, we keep the same shapes for input and output nodes of each layer. In the first layer, we adopt different input settings for single text tasks such as sentiment classification and text pair tasks such as textual entailment. As shown in Figure~\ref{fig:search-space}, the inputs $c_{k-2}$ and $c_{k-1}$ are set as the same text input for single text tasks, and set as two input texts for text pair tasks. This setting helps to explore self-attention or self-interactions for single text task, and pair-wise cross-interactions for text pair tasks.

For the candidate operations in the cell, we adopt lightweight CNN-based operations as they are effective in NLP tasks \cite{bai2018empirical}, and CNN-based operations have shown inference speed superiority over RNN-based models and self-attention based models \cite{shen2018bidirectional} due to the fact that they are parallel-friendly operations. 
Specifically, the candidates operation set $\mathcal{O}$ include $convolution$, $pooling$, $skip$ (identity) connection and $zero$ (discard) operation.
For $convolution$ operations, both $1$D standard convolution and dilated convolution with kernel size $\{3, 5, 7\}$ are included, among which the dilated convolution can be used to enhance the capability of capturing long-dependency information and each convolution is applied as a Relu-Conv-BatchNorm structure. 
$Pooling$ operations include averaging pooling and max pooling with kernel size $3$.
The $skip$ and $zero$ operations are used to build residual connection and discard operation respectively, which are helpful to reduce network redundancy.
Besides, for both convolution and pooling operations, we apply the ``SAME" padding
to make the output length as the same as the input.

 \begin{table*}[tp]
 \centering
 \small
 \begin{tabular}{c c c c c c c c c c} 
 \toprule
 \multirow{2}{*}{Method} &  \multirow{2}{*}{\# Params} & Inference & \multirow{2}{*}{SST-2} & \multirow{2}{*}{MRPC} &  \multirow{2}{*}{QQP} &  \multirow{2}{*}{MNLI} &  \multirow{2}{*}{QNLI} &  \multirow{2}{*}{RTE} &  \multirow{2}{*}{Average}\\ 
  &   & Speedup & & & & & & &  \\
  \midrule 
  BERT$_{12}$ & \multirow{2}*{109M}  & \multirow{2}*{1x} & 93.5 & 88.9 & 71.2 & 84.6 & 90.5 & 66.4 & 82.5\\ 
  BERT$_{12}$-T  & & & 93.3 & 88.7 & 71.1 & 84.8 & 90.4 & 66.1 & 82.4\\
  \midrule
 BERT$_6$-PKD & 67.0M & 1.9x & \underline{92.0} & 85.0 &    \underline{70.7} &    81.5 &    \textbf{89.0} &    \textbf{65.5} & \textbf{80.6}\\  
 BERT$_3$-PKD & 45.7M & 3.7x & {87.5} & {80.7} & {68.1} & {76.7} & {84.7} & {58.2} & 76.0 \\ 
 DistilBERT$_4$ & 52.2M & 3.0x & 91.4 & 82.4 &    68.5 &    78.9 &    85.2 &    54.1 & 76.8\\  
 TinyBert$_4$ & 14.5M & \underline{9.4x} & \textbf{92.6} & \textbf{86.4} &    \textbf{71.3} &    \textbf{82.5} &    \underline{87.7} &    {62.9} & \textbf{80.6}\\  
 BiLSTM$_{SOFT}$ & \underline{10.1M} & 7.6x & {90.7} & - &    {68.2} &    {73.0} &    {-} &    {-} & -\\
  \midrule
 AdaBERT & \textbf{6.4M $\sim$ 9.5M}  & \textbf{12.7x $\sim$ 29.3x} & {91.8} & \underline{85.1}  & \underline{70.7} & \underline{81.6} & {86.8} & \underline{64.4} & \underline{80.1}\\ 
 \bottomrule
 \end{tabular}
 \caption{The compression results including model efficiency and accuracy from the GLUE test server, and the MNLI result is evaluated for matched-accuracy (MNLI-m). 
 BERT$_{12}$ indicates the results of the fine-tuned BERT-base from \protect\cite{devlin2019bert} and BERT$_{12}$-T indicates the results of the fine-tuned BERT-base in our implementation.
 The results of BERT-PKD are from \protect\cite{sun2019patient}, the results of DistilBERT$_4$ and TinyBERT$_4$ are from \protect\cite{jiao2019tinybert}, and the results of BiLSTM$_{SOFT}$ is from \protect\cite{tang2019distilling}.
 The number of model parameters includes the embedding size, and the inference time is tested with a batch size of $128$ over $50,000$ samples. 
 The bold numbers and underlined numbers indicate the best and the second-best performance respectively.}
 \label{tab-overall}
 \end{table*}

\begin{table}[tp]
\small
\centering
\begin{tabular}{c c c c} 
\toprule
\multirow{2}{*}{Task} & \multirow{2}{*}{$K$} & \multirow{2}{*}{\# Params} & Inference \\
& & & Speedup \\
\midrule
SST-2 & 3 & 6.4M  & 29.3x \\
MRPC  & 4 & 7.5M  & 19.2x \\
QQP   & 5 & 8.2M  & 16.4x \\
MNLI  & 7 & 9.5M  & 12.7x \\
QNLI  & 5 & 7.9M  & 18.1x \\
RTE   & 6 & 8.6M  & 15.5x \\
\bottomrule
\end{tabular}
\caption{The number of layers, parameters and inference speedups of searched structures by AdaBERT for different tasks.}
\label{tab-inference}
\end{table}
\subsubsection{Search Algorithm}
Directly optimizing the overall loss $\mathcal{L}$ in Equation (\ref{equ:overall-loss}) by brute-force enumeration of all the candidate operations is impossible, due to the huge search space with combinatorial operations and the time-consuming training on its $w_{\alpha}$ w.r.t. $\alpha$. 
Here we solve this problem by modeling searched architecture $\{K, o_{i,j}\}$ as discrete variables that obey discrete probability distributions $P_K = [\theta^K_1, \dots, \theta^K_{K_{max}}]$ and $P_o = [\theta^o_1, \dots, \theta^o_{|\mathcal{O}|}]$. $K$ and $o_{i,j}$ are thus modeled as one-hot variables and sampled from layer range $[1,K_{max}]$ and candidate operation set $\mathcal{O}$ respectively.
However, $\mathcal{L}$ is non-differentiable as the discrete sampling process makes the gradients cannot propagate back to the learnable parameters $P_K$ and $P_o$.
Inspired by \cite{xie2019snas,wu2019fbnet}, we leverage Gumbel Softmax technique \cite{jang2017categorical,chris2017concrete} to relax the categorical samples into continuous sample vectors $y^K \in R^{K_{max}}$ and $y^o \in R^{|\mathcal{O}|}$ as:
\begin{equation}
\label{eqn:gumbel_softmax}
\begin{aligned}
    y_{i}^K  &= \frac{\exp[(log(\theta_{i}^K) + g_{i})/\tau]}{\sum_{j=1}^{K_{max}} \exp[(log(\theta_{j}^K) + g_{j})/\tau]}, \\
    y_{i}^o  &= \frac{\exp[(log(\theta_{i}^o) + g_{i})/\tau]}{\sum_{j=1}^{|\mathcal{O}|} \exp[(log(\theta_{j}^o) + g_{j})/\tau]},
\end{aligned}
\end{equation}
where $g_{i}$ is a random noise drawn from Gumbel(0, 1) distribution, $\tau$ is the temperature to control the degree of approximating Gumbel-Softmax to \textit{argmax}, i.e., as $\tau$ approaches 0, the samples become one-hot. 
By this way, $y^K$ and $y^o$ are differentiable proxy variables to discrete samples, and then we can efficiently optimize $\mathcal{L}$ directly using gradient information.
Specifically, we use one-hot sample vectors $argmax(y^K)$ and $argmax(y^o)$ in the forward stage while use continuous $y^K$ and $y^o$ in the back-propagation stage, which is called Straight-Through Estimator \cite{bengio2013estimating} and makes the forward process in training consistent with testing.

Note that the optimization of the overall loss $\mathcal{L}$ can be regarded as learning a parent graph defined by the search space $\alpha$ as described in Section \ref{sec:space}, whose weights of candidate operations $w_{\alpha}$ and architecture distribution $P_{\alpha}$ are trained simultaneously. In the training stage, the randomness introduced by $P_{\alpha}$ enhances the exploring for suitable contextual encoders that mimic task-specific teacher $BERT_t$ under resource constraints. 
After the training of the parent graph, we can derive an efficient and task-adaptive child graph by applying $argmax$ on $P_{\alpha}$ as the compressed model.
Also note that in $\mathcal{L}$, the knowledge distillation loss $\mathcal{L}_{KD}$  provides regularization for the architecture sampling on $P_{\alpha}$, while the efficiency-aware loss $\mathcal{L}_E$ promotes sparse structures that make the model compact and efficient.

\section{Experiments}

\subsection{Setup}
\paragraph{Datasets.}
We evaluate the proposed AdaBERT method on six datasets from GLUE \cite{wang2018glue} benchmark.
Specifically, we consider three types of NLP tasks, namely sentiment classification, semantic equivalence classification, and entailment recognition.
\textit{SST-2}
is adopted for sentiment classification, whose goal is to label movie reviews as positive or negative.
\textit{MRPC}
and 
\textit{QQP}
are adopted for semantic equivalence classification, whose sentence pairs are extracted from news sources and online website respectively.
\textit{MNLI}, \textit{QNLI} and \textit{RTE}
are adopted for textual entailment recognition, whose premise-hypothesis pairs vary in domains and scales.

\paragraph{Baselines.}
We compare the proposed AdaBERT method with several state-of-the-art BERT compression methods including BERT-PKD~\cite{sun2019patient}, DistilBERT~\cite{sanh2019distilbert}, TinyBERT~\cite{jiao2019tinybert} and BiLSTM$_{SOFT}$~\cite{tang2019distilling}.
Since our approach searches architecture from a space including the number of network layers $K$, here we also compare with several different versions of these baselines with the different number of layers for a comprehensive comparison. 

\paragraph{AdaBERT Setup.}
We fine-tune the BERT-base model~\cite{devlin2019bert} on the six adopted datasets respectively as teacher models for knowledge distillation.
For input text, following~\cite{lan2019albert}, we factorize the WordPiece embedding of $BERT_t$ into smaller embeddings whose dimension is $128$ and set the max input length as $128$.
For the optimization of operation parameters, we adopt SGD with momentum as $0.9$ and learning rate from 2e-2 to 5e-4 scheduled by cosine annealing.
For the optimization of architecture distribution parameters $P_\alpha$, we use Adam with a learning rate of 3e-4 and weight decay of 1e-3.
For AdaBERT, we set $\gamma=0.8$, $\beta=4$, $T=1$, inner node $N=3$ and search layer $K_{max}=8$. 
We search $P_\alpha$ for 80 epochs and derive the searched structure with its trained operation weights.

 \begin{table*}[htp]
 \centering
 \small
 \begin{tabular}{|c| c| c c| c c c|} 
 \hline
Structure $\backslash$ Task & SST-2 & MRPC & QQP & MNLI & QNLI & RTE \\ 
 \hline
 AdaBERT-SST-2 & \textbf{91.9} & 78.1  &    58.6 &  64.0 & 74.1 &  53.8     \\  
 \hline
 AdaBERT-MRPC  & 81.5  & \textbf{84.7} & {68.9} & {75.9} & {82.2} & {60.3} \\ 
 AdaBERT-QQP   & 81.9 & 84.1 &    \textbf{70.5} &    76.3 &  82.5     &    60.5 \\  
 \hline
 AdaBERT-MNLI  & 82.1  & 81.5 &    66.8 & \textbf{81.3} & 86.1     &63.2     \\  
 AdaBERT-QNLI  & 81.6  & 82.3 &    67.7 & 79.2     &    \textbf{87.2} &    62.9 \\  
 AdaBERT-RTE   &  82.9 & 81.1 &    66.5 & 79.8     &    86.0 &    \textbf{64.1} \\  
 \hline
 Random   & 80.4 $\pm$ 4.3 & 79.2 $\pm$ 2.8 &  61.8 $\pm$ 4.9  & 69.7 $\pm$ 6.7     &  78.2 $\pm$ 5.5   &  55.3 $\pm$ 4.1   \\  
  \hline
 \end{tabular}

 \caption{Accuracy comparison on the dev sets with the searched compression structures applying to different tasks. For Random, 5-times averaging results with standard deviations are reported. 
 }
 \label{tab-structure-adapt}
 \end{table*}

\subsection{Overall Results}
\subsubsection{Compression Results}
\label{exp:overall-res}
The compression results on the six adopted datasets, including parameter size, inference speedup and classification accuracy, are summarized in Table \ref{tab-overall}. Detailed results of AdaBERT method for different tasks are reported in Table \ref{tab-inference}.

Overall speaking, on all the evaluated datasets, the proposed AdaBERT method achieves significant efficiency improvement while maintaining comparable performance. 
Compared to the BERT$_{12}$-T, the compressed models are 11.5x to 17.0x smaller in parameter size and 12.7x to 29.3x faster in inference speed with an average performance degradation of $2.79\%$. This demonstrates the effectiveness of AdaBERT to compress BERT into task-dependent small models.

Comparing with different Transformers-based compression baselines, 
the proposed AdaBERT method is 1.35x to 3.12x faster than the fastest baseline, TinyBERT$_4$, and achieves comparable performance with the two baselines that have the best averaged accuracy, BERT$_6$-PKD and TinyBERT$_4$.
Further, as shown in Table \ref{tab-inference}, AdaBERT searches suitable layers and structures for different tasks, e.g., the searched structure for SST-2 task is lightweight since this task is relatively easy and a low model capacity is enough to mimic task-useful knowledge from the original BERT. This observation confirms that AdaBERT can automatically search small compressed models that adapt to downstream tasks.

Comparing with another structure-heterogeneous method, BiLSTM$_{SOFT}$, AdaBERT searches CNN-based models and achieves much better improvements, especially on the MNLI dataset.
This is because AdaBERT searches different models for different downstream tasks (as Table \ref{tab-inference} shows), and adopts a flexible searching manner to find suitable structures for different tasks while BiLSTM$_{SOFT}$ uses a Siamese structure for all different tasks. This shows the flexibility of AdaBERT to derive task-oriented compressed models for different tasks, and we will investigate more about this in the following part.

\subsubsection{Adaptiveness Study}
\label{exp:structure-study}
\paragraph{Cross-Task Validation.}
In order to further examine the adaptiveness of searched structures by AdaBERT, we apply the searched compression model structures across different downstream tasks. For example, the searched structure for task SST-2 (we denote this searched structure as AdaBERT-SST-2) is applied to all different tasks. For such cross-task validation, we randomly initialize the weights of each searched structure and re-train its weights using corresponding training data to ensure a fair comparison. The results of cross-task validation is summarized in Table \ref{tab-structure-adapt}.

From Table \ref{tab-structure-adapt}, we can observe that:
The searched structures achieve the best performance on their original target tasks compared with other tasks, in other words, the performance numbers along the diagonal line of this table are the best. 
Further, the performance degradation is quite significant across different kinds of tasks (for example, applying the searched structures of sentiment classification tasks to entailment recognition task, or vice verse), 
while the performance degradations within the same kind of tasks (for example, MRPC and QQP for semantic equivalence classification) are relatively small, since they have the same input format (i.e., a pair of sentences) and similar targets.
All these observations verify that AdaBERT can search task-adaptive structures for different tasks with the guidance of task-specific knowledge.

We also conduct another set of experiment: for each task, we randomly sample a model structure without searching, and then train such structures on the corresponding datasets.
From the last row of Table \ref{tab-structure-adapt}, we can see that the randomly sampled structures perform worse than the searched structures and their performances are not stable. This shows the necessity of the proposed adaptive structure searching.

\begin{figure}[t]
\centering
\subfloat [Sentiment Classification, SST-2]{
\includegraphics[height=0.6in,width=3in]{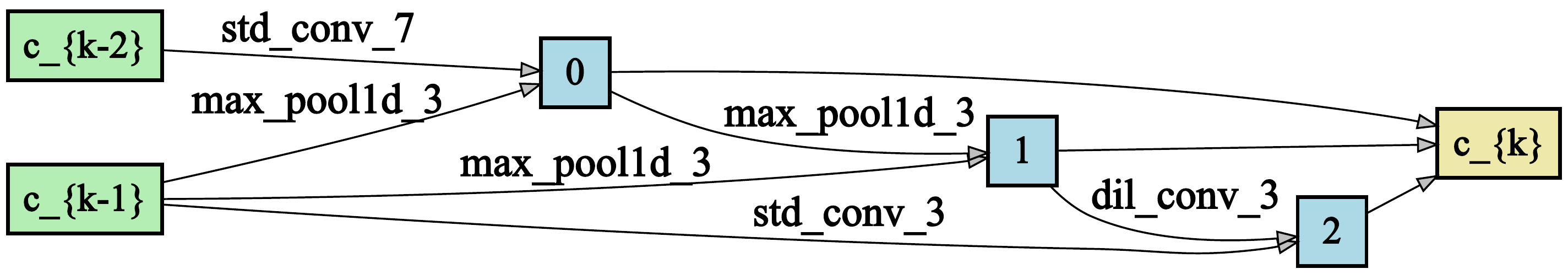}
\label{fig:structure1}
}
\vfil
\subfloat [Equivalence Classification, MRPC]{
\includegraphics[height=0.63in,width=2.7in]{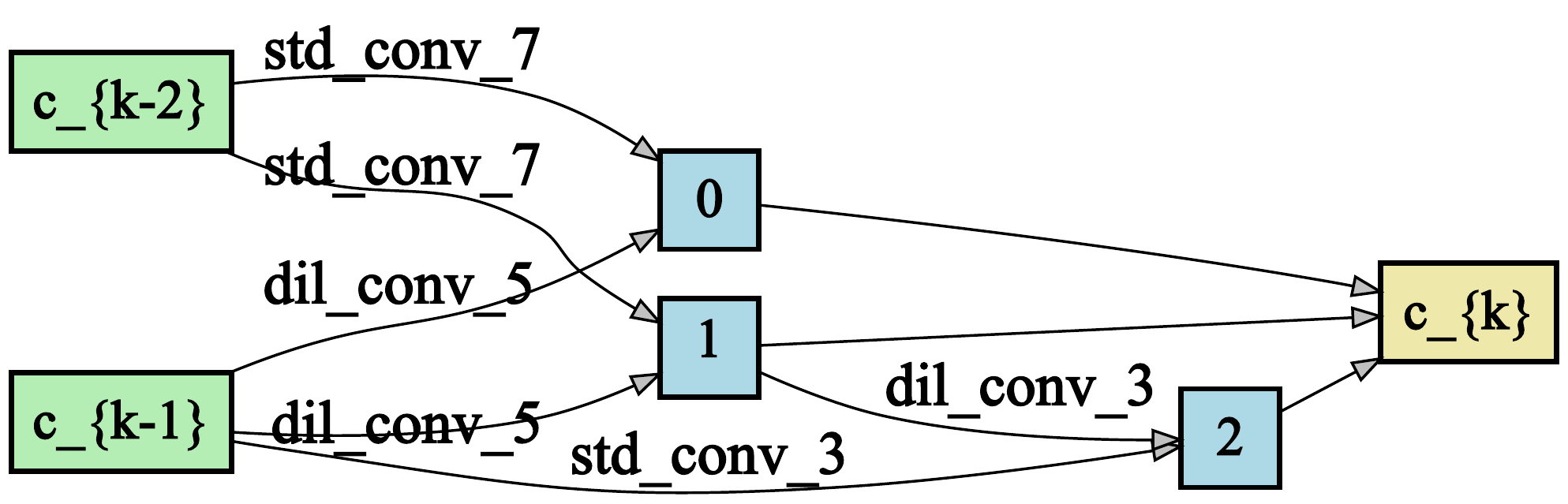}
\label{fig:structure2}
}
\vfil
\subfloat [Entailment Recognition, QNLI]{
\includegraphics[height=0.73in,width=2.8in]{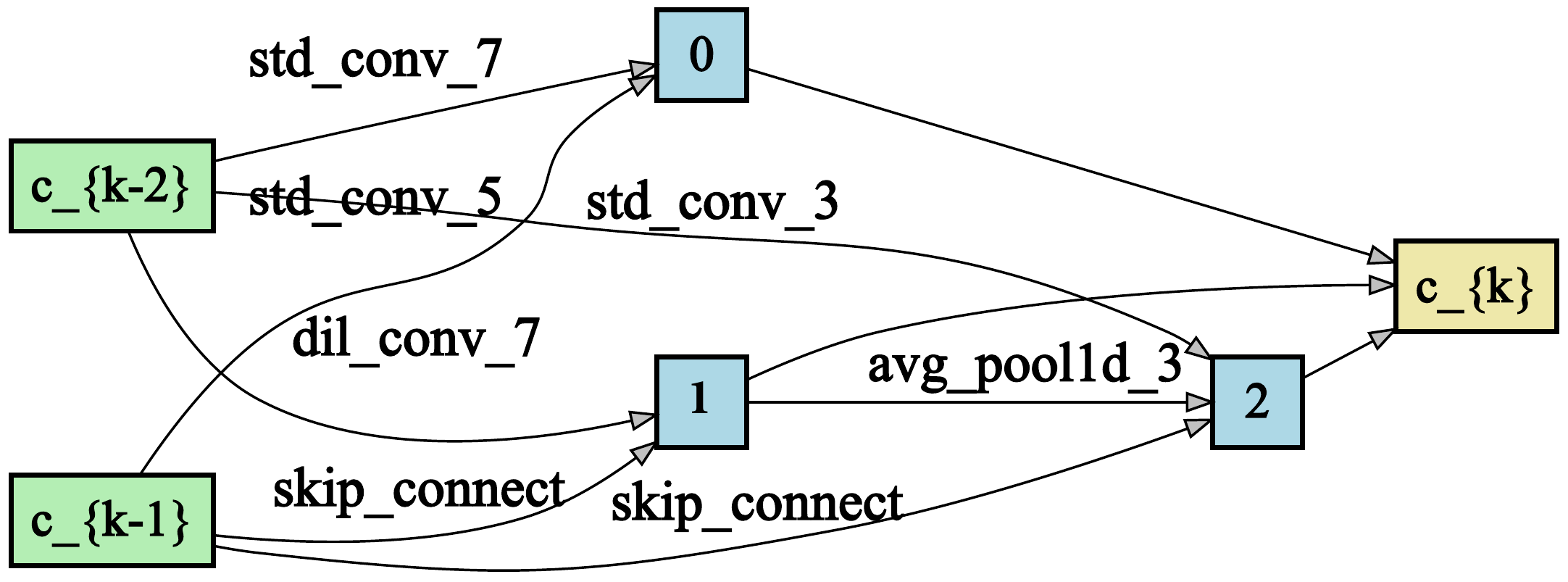}
\label{fig:structure3}
}
\caption{The searched basic cells for three kinds of NLP tasks.}
\label{fig:structure}
\end{figure}

\paragraph{Architecture Study.}
In order to examine the adaptiveness of searched structures, we also visualize the basic cells of searched structures for different tasks in Figure \ref{fig:structure}. 

By comparing the structure searched for single-text task such as sentiment classification (Figure~\ref{fig:structure1}) with the one for sentiment equivalence classification task (Figure~\ref{fig:structure2}), we can find that the former searched structure has more aggregation operations (\textit{max\_pool1d\_3}) and smaller feature filters (\textit{std\_conv\_3} and \textit{dil\_conv\_3}) since encoding local features are good enough for the binary sentiment classification task, while the latter searched structure has more interactions between the two input nodes as it deals with text pairs. 

For text-pair tasks, compared with sentiment equivalence classification task (Figure \ref{fig:structure2}), the searched structure for the entailment recognition task (Figure \ref{fig:structure3}) has more diverse operations such as \textit{avg\_pool\_3} and \textit{skip\_connect}, and more early interactions among all the three subsequent nodes (node $0$, $1$ and $2$).
This may be justified by the fact that textual entailment requires different degrees of reasoning and thus the searched structure has more complex and diverse interactions.

The above comparisons among the searched structures for different tasks confirm that AdaBERT can search task-adaptive structure for BERT compression. 
Next, we conduct ablation studies to examine how knowledge losses and efficiency-ware loss affect the performance of AdaBERT.

\begin{table}[bt]
\small
\centering
\begin{tabular}{p{1.3cm} c c c c c c c} 
\toprule
 & SST-2 & MRPC & QQP & MNLI & QNLI & RTE \\ 
\midrule
 Base-KD    & {86.6} & {77.2}  & 63.9 & 75.2 & {82.0}    & {56.7} \\
~ + Probe    & {88.4} & {78.7}  & 67.3 & 77.8 &{83.3} & {58.1} \\ 
~ + DA      & {91.4} & {83.9}  & 70.1 & 80.7&{86.5} & {63.2} \\
 +$\mathcal{L}_{CE}$(All) & {91.9}& {84.7} &70.5 & 81.3 & {87.2} & {64.1} \\
 \bottomrule
\end{tabular}
\caption{The effect of knowledge loss terms. }
\label{tab-kg-ablation}
\end{table}

\begin{table}[bt]
\small
\centering
\begin{tabular}{p{0.54cm} c c c c c c} 
\toprule
 & SST-2 & MRPC & QQP & MNLI& QNLI & RTE \\ 
\midrule
\multirow{2}*{$\beta$ = 0}  & {91.8} & {84.5} & 70.3 & 81.2 &  87.1 & {63.9} \\
   & (7.5M) & (7.8M) & (8.6M)  & (9.6M) & (8.3M) &(9.1M) \\
\midrule
\multirow{2}*{$\beta$ = 4}  & \textbf{91.9} & \textbf{84.7} &  \textbf{70.5}  & \textbf{81.3} &\textbf{87.2} & \textbf{64.1} \\
   & (6.4M) & (7.5M) & (8.2M) & (9.5M) &(7.9M)& (8.6M) \\
\midrule
\multirow{2}*{$\beta$ = 8} & {91.3} & {84.2} & 68.9 & 79.6 & 86.4& {63.3} \\
   & (5.3M) & (6.4M) & (7.3M) & (7.1M) &  (7.2M)& (7.8M) \\
\bottomrule
\end{tabular}

\caption{The effect of efficiency loss term. }

\label{tab-l3-ablation}
\end{table}
\subsection{Ablation Study}
\label{exp:ablation-res}
\subsubsection{Knowledge Losses}

We evaluate the effect of knowledge distillation ($\mathcal{L_{KD}}$) and the supervised label knowledge ($L_{CE}$) by conducting experiments on different tasks. The results are shown in Table \ref{tab-kg-ablation}.

The Base-KD (line 1 in Table \ref{tab-kg-ablation}) is a naive knowledge distillation version in which only the logits of the last layer are distilled without considering hidden layer knowledge and supervised label knowledge. 
By incorporating the probe models, the performance (line 2 in Table \ref{tab-kg-ablation}) is consistently improved, indicating the benefits from hierarchically decomposed task-oriented knowledge.
We then leverage Data Augmentation (DA) to enrich task-oriented knowledge and this technique also improves performance for all tasks, especially for tasks that have a limited scale of data (i.e., MRPC and RTE).
DA is also adopted in existing KD-based compression studies \cite{tang2019distilling,jiao2019tinybert}.

When taking the supervised label knowledge ($\mathcal{L}_{CE}$) into consideration, the performance is further boosted, showing that this term is also important for AdaBERT by providing focused search hints.

\subsubsection{Efficiency-aware Loss}
Last, we test the effect of efficiency-aware loss $\mathcal{L}_E$ by varying its corresponding coefficient, including the standard case ($\beta=4$), without efficiency constraint ($\beta=0$), and strong efficiency constraint ($\beta=8$).
The model performance and corresponding model size are reported in Table \ref{tab-l3-ablation}.

On the one hand, removing the efficiency-aware loss ($\beta=0$) leads to the increase in model parameter size, on the other hand, a more aggressive efficiency preference ($\beta=8$) results in the small model size but degraded performance, since a large $\beta$ encourages the compressed model to adopt more lightweight operations such as $zero$ and $skip$ which hurt the performance.
A moderate efficiency constraint ($\beta=4$) provides a regularization,  guiding the AdaBERT method to achieve a trade-off between the small parameter size and the good performance.

\section{Related Work}
Existing efforts to compress BERT model can be broadly categorized into four lines: knowledge distillation, parameter sharing, pruning and quantization. 

For knowledge distillation based methods, in \cite{tang2019distilling}, BERT is distilled into a simple BiLSTM and achieves comparable results with ELMo. 
A dual distillation is proposed to reduce the vocabulary size and the embedding size in \cite{zhao2019extreme}.
PKD-BERT \cite{sun2019patient} and DistilBERT \cite{sanh2019distilbert} distill BERT into shallow Transformers in the fine-tune stage and the pre-train stage respectively.   
TinyBERT \cite{jiao2019tinybert} further distills BERT with a two-stage knowledge distillation.
For parameter sharing based methods, the multi-head attention is compressed into a tensorized Transformer in \cite{ma2019tensorized}. 
AlBERT \cite{lan2019albert} leverages cross-layer parameter sharing to speed up the training.
Different from these existing methods, the proposed AdaBERT incorporates both task-oriented knowledge distillation and efficiency factor, and automatically compresses BERT into task-adaptive small structures instead of a task-independent structure in existing methods.

For pruning and quantization based methods, \cite{Michel2019aresixteen} and \cite{wang2019structured} prune attention heads and  weight matrices respectively.
Q8BERT \cite{zafrir2019q8bert} quantizes matrix multiplication operations into 8-bit operations, while Q-BERT \cite{shen2019qbert} quantizes BERT with Hessian based mix-precision.
These methods and the proposed AdaBERT compress BERT from different aspects that are complementary, that is, one can first distill BERT into a small model, and then prune or quantize the small model.

\section{Conclusion}
In this work, motivated by the strong need to compress BERT into small and fast models, we propose AdaBERT, an effective and efficient model that adaptively compresses BERT for various downstream tasks.
By leveraging Neural Architecture Search, we incorporate two kinds of losses, task-useful knowledge distillation loss and model efficiency-aware loss, such that the task-suitable structures of compressed BERT can be automatically and efficiently found using gradient information.
We evaluate the proposed AdaBERT on six datasets involving three kinds of NLP tasks. 
Extensive experiments demonstrate that AdaBERT can achieve comparable performance while significant efficiency improvements, and find task-suitable models for different downstream tasks.
\balance

\newpage
{
\bibliography{main}
}
\bibliographystyle{ijcai_natbib}

\end{document}